\documentclass{article}
\usepackage[preprint]{neurips_2024}

\author{
Nanbeige LLM Lab, Boss Zhipin
}

\usepackage{neurips_2024}

\usepackage{booktabs}
\usepackage{makecell}  

\usepackage[utf8]{inputenc} 
\usepackage[T1]{fontenc}    
\usepackage{hyperref}       
\usepackage{url}            
\usepackage{booktabs}       
\usepackage{amsfonts}       
\usepackage{nicefrac}       
\usepackage{microtype}      
\usepackage{xcolor}         
\usepackage{minted}

\usepackage{listings}
\usepackage{xcolor}  

\usepackage{tcolorbox}
\tcbuselibrary{listings, skins}
\tcbuselibrary{breakable}
\usepackage{algorithm}
\usepackage{algpseudocode}
\usepackage{amsmath}
\usepackage{url}
\usepackage{graphicx}
\usepackage{amsmath}
\usepackage{amsthm}
\usepackage{booktabs}
\usepackage{color}
\usepackage{xcolor}
\usepackage{xspace}
\usepackage[misc]{ifsym}

\usepackage[ruled, vlined, nofillcomment, linesnumbered, algo2e]{algorithm2e}

\usepackage{booktabs}
\usepackage{multirow}
\usepackage{balance}
\usepackage{enumitem}
\usepackage{stfloats}
\usepackage{diagbox}
\usepackage{fancyhdr}
\usepackage{stfloats}
\definecolor{result_color}{RGB}{250,250,210}
\usepackage{bm}

\newcommand{\ignore}[1]{}
\newcommand{\paratitle}[1]
{\vspace{1.5ex}\noindent\textbf{#1}}

\title{\large{Nanbeige4.1-3B: A Small General Model that Reasons, Aligns, and Acts}}

\begin{document}
\noindent\includegraphics[height=0.8cm]{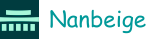}

\maketitle

\begin{abstract}
We present Nanbeige4.1-3B, a unified generalist language model that simultaneously achieves strong agentic behavior, code generation, and general reasoning with only 3B parameters. 
To the best of our knowledge, it is the first open-source small language model (SLM) to achieve such versatility in a single model.
To improve reasoning and preference alignment, we combine point-wise and pair-wise reward modeling, ensuring high-quality, human-aligned responses.
For code generation, we design complexity-aware rewards in Reinforcement Learning, optimizing both correctness and efficiency.
In deep search, we perform complex data synthesis and incorporate turn-level supervision during training. This enables stable long-horizon tool interactions, allowing Nanbeige4.1-3B to reliably execute up to 600 tool-call turns for complex problem-solving.
Extensive experimental results show that Nanbeige4.1-3B significantly outperforms prior models of similar scale, such as Nanbeige4-3B-2511 and Qwen3-4B, even achieving superior performance compared to much larger models, such as Qwen3-30B-A3B.
Our results demonstrate that small models can achieve both broad competence and strong specialization simultaneously, redefining the potential of 3B parameter models. The model checkpoint is available at~\url{https://huggingface.co/Nanbeige/Nanbeige4.1-3B}

\end{abstract}

\begin{figure}[H]
 	\centering
 	\includegraphics[width=0.9\textwidth]{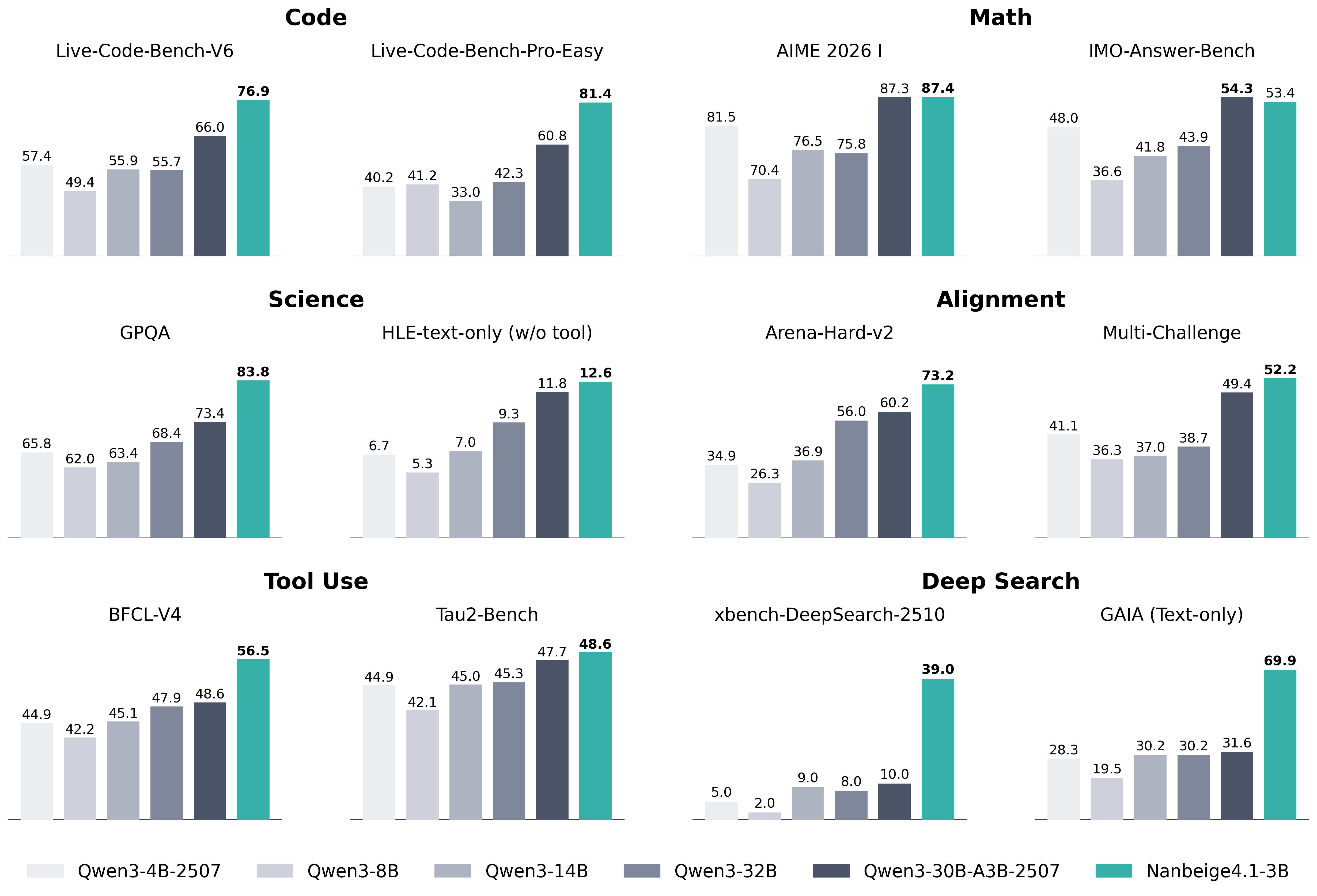}
	\caption{\textcolor{black}{Performance of Nanbeige4-3B-Thinking vs. Qwen model series.
  }}
	\label{fig:thinking_performance}
 \end{figure}

\section{Introduction}\label{sec:intro}

Recent advances in small language models~(SLMs) have demonstrated that compact models can achieve impressive performance on specialized tasks such as mathematical reasoning or code generation ~\cite{yang2025nanbeige4, yan2026distribution}. 
Despite the small size, these models perform competitively by leveraging innovations in architecture, training data, and training algorithm, making them highly effective in practical applications.

However, most existing SLMs exhibit fragmented capabilities: reasoning-focused models often struggle with long-horizon interactions (e.g., deep search) \cite{xu2025tiny}, while code or agent specialized models typically lack robust general reasoning abilities, such as creative writing or human preference alignment \cite{zhu2026re}. Consequently, building a truly unified generalist model at the 3B scale remains an open challenge
and a practical question is raised: \textbf{How far can a 3B model be pushed as a generalist without compromising its existing strengths?}

In this work, we present \textbf{Nanbeige4.1-3B}, a generic small language model that is remarkably powerful in reasoning, coding, and agentic behaviors,
Built upon Nanbeige4-3B \cite{yang2025nanbeige4}. For general reasoning, we extend our previous pair-wise preference modeling by incorporating point-wise reward signals, ensuring that responses are both strong in isolation and preferred under direct comparison.
For code generation, we move beyond correctness as the sole objective and explicitly reward algorithmic efficiency, encouraging solutions that are not only functionally correct but also computationally efficient.
For agentic behavior, we emphasize long-horizon planning. We construct high-quality training data through Wiki-graph random walks and define rewards at both the interaction (turn) and full-trajectory levels, allowing the model to receive credit for planning and execution over hundreds of steps.
Throughout training, we employ careful SFT data mixing and multi-stage reinforcement learning \cite{wang2025nemotron} to maintain balance across these domains.

Through the aforementioned approach, we have trained a model with an exceptionally broad and stable capability profile.
Across reasoning and coding tasks, Nanbeige4.1-3B consistently surpasses existing open-source SLMs (e.g., Qwen3-4B and Qwen3-8B) and shows clear improvements over Nanbeige4-3B-Thinking-2511. 
More notably, Nanbeige4.1-3B exhibits deep-search and long-horizon agentic behavior rarely observed in general-purpose small language models. 
While models such as Qwen3-4B and Qwen3-8B fail to sustain meaningful exploration beyond a few turns, Nanbeige4.1-3B reliably solves complex search-oriented tasks through extended tool interaction. 
In this regime, a generalist 3B model attains deep-search capability comparable to that of specialized search-oriented models at the tens-of-billions scale, and approaches the search performance of 100B+ general-purpose large models. These results indicate that long-horizon agency can be achieved in compact generalist models when training objectives and credit assignment are properly aligned.

We open-source Nanbeige4.1-3B to support research into efficient, agent-capable language models. Beyond the checkpoint itself, we hope this release contributes to the community's understanding of how to jointly train reasoning, coding, and long-horizon behaviors under strict capacity constraints.

\section{Methods}
In this section, we present the methodology used to equip a compact 3B model with broad generalist capabilities. We first describe how we individually optimize the model for general reasoning, long-horizon search (agentic behavior), and code generation, focusing on the design of training signals and reward structures tailored to each capability. We then detail how these heterogeneous objectives are integrated through data mixing and multi-stage training, enabling the model to retain domain-specific strengths while emerging as a unified general model under strict capacity constraints.

\subsection{General Abilities}
In this section, we elaborate on the optimization of Nanbeige4.1's general capabilities. Our improvements focus on two key aspects: refining the data construction strategies in the SFT phase, and upgrading the General RL training paradigm through a progressive integration of Point-wise and Pair-wise reinforcement learning.

\subsubsection{SFT}
Nanbeige4.1-3B is built upon the Nanbeige4-3B-Base \cite{yang2025nanbeige4} with an enhanced SFT recipe, focusing on data distribution, length scaling, and training data quality.

First, we redesign the SFT data mixture. Compared to the previous Nanbeige4-3B-2511 version, we increase the proportion of code-related data and introduce a higher ratio of challenging problems in mathematics and general domains. This shift encourages stronger reasoning depth and improves robustness on difficult benchmarks.

Second, we extend the context length beyond the previous two-stage curriculum (32k $\rightarrow$ 64k) by introducing a third stage at 256k tokens to better support complex reasoning and long-horizon scenarios. In the final 256k stage, we adopt a specialized data mixture designed to strengthen agentic and reasoning capabilities, consisting of code (27\%), deep-Search (26\%), STEM (23\%), tool-use (13\%), and general domains (10\%).

Third, we further optimize the Solution Refinement and Chain-of-Thought (CoT) Reconstruction framework originally introduced in Nanbeige4-3B-2511. Specifically, we scale up the number of refinement iterations in the Solution Refinement loop, allowing stronger critique–revision cycles to produce higher-quality final solutions. In addition, we train a more capable CoT Reconstruction model to generate cleaner and more faithful reasoning traces from refined answers.

As shown in Table~\ref{tab:phase1_sft}, these improvements result in substantial gains across benchmarks. Nanbeige4.1-3B-SFT demonstrates consistent improvements in coding, mathematics, and alignment metrics, laying a stronger foundation for subsequent reinforcement learning stages.

\begin{table}[H] 
\centering
\caption{Performance Uplift from Nanbeige4-3B-SFT to Nanbeige4.1-3B-SFT}
\label{tab:phase1_sft}
\small 
\resizebox{0.95 \textwidth}{!}{%
\begin{tabular*}{\textwidth}{@{\extracolsep{\fill}}llccc}
\toprule
\textbf{Domain} & \textbf{Benchmark} & \textbf{Nanbeige4-3B-SFT} & \textbf{Nanbeige4.1-3B-SFT} & \textbf{$\Delta$} \\ 
\midrule
\multirow{2}{*}{\textbf{Code}}  & LCB V6 & 45.5 & \textbf{62.0} & +16.5 \\
& LCB Pro Medium  & 1.8 & \textbf{22.8} & +21.0 \\
\midrule
\multirow{2}{*}{\textbf{Math}} 
 & Hmmt Nov & 60.7 & \textbf{74.3} & +13.6 \\
 & Imo-Answer-Bench & 34.8 & \textbf{48.9} & +14.1 \\ 
 \midrule
\multirow{2}{*}{\textbf{Alignment}} & Arena-Hard V2 & 45.5 & \textbf{60.2} & +14.7 \\
 & Multi-Challenge & 42.6 & \textbf{44.4} & +1.8 \\ 
\bottomrule
\end{tabular*}
}
\end{table}

\subsubsection{Point-wise RL}
After SFT, we still observe some degradation issues in Nanbeige4.1-3B-SFT, such as repetition and redundant thinking, which have also been reported in prior work~\cite{feng2026paceprefixprotecteddifficultyawarecompression}. To address these issues and establish a more stable foundation for RL, we introduce a point-wise reinforcement learning stage.

We train a general reward model to evaluate rollout responses, following prior work~\cite{wang2025nemotron}. The model is trained on curated large-scale human preference data. We found that the trained reward model naturally suppresses overly redundant, repetitive, and low-readability answers. We then perform GRPO \cite{shao2024deepseekmath} to optimize Nanbeige4.1-3B-SFT, sampling 8 rollouts per prompt and using the general reward model to score each response as the training signal.

With point-wise RL, we significantly reduce formatting errors and redundant reasoning. On LiveCodeBench-v6, the point-wise RL greatly improves length stability, reducing overlong truncation from 5.27\% to 0.38\%. As shown in Table ~\ref{tab:phase2_rl}, it also advances Arena-Hard V2 from 60.2 to 66.6, with the hard-prompt subset improving from 46.1 to 54.1. These gains are reflected in more consistent, well-structured outputs and high-quality code presentation.

\subsubsection{Pair-wise RL}
Although the Point-wise RL training provides effective alignment signals, the amount of high-quality preference data is limited, which constrains further improvement. To address this, we introduce Pair-wise RL to fully leverage preference information from strong–weak model comparisons and further enhance model performance.

We train a pair-wise reward model on paired comparison data spanning code generation and LMArena-style conversations (single-turn and multi-turn). We generate response pairs with a strong model and a weak model, then apply the same checklist filtering strategy as Nanbeige4 to derive reliable win–loss labels~\cite{yang2025nanbeige4}. Following the framework~\cite{xu2025unified}, we mitigate position bias by adding a swap-consistency regularizer, defined as the mean squared error between the predicted reward difference for a response pair and the negated reward difference for the swapped pair.

We then run Pair-wise RL by formulating the reward as a binary outcome based on whether the generated rollout outperforms the reference answer. For multi-turn scenarios, we concatenate the full dialogue history into the pairwise reward model's input. 

As shown in ~\ref{tab:phase2_rl}, Nanbeige4.1-3B achieves comprehensive performance breakthroughs after Pair-wise RL. By deeply exploiting contextual information in multi-turn dialogues, Alignment metrics show significant gains, with Multi-Challenge increasing from 47.72 to 55.14. Additionally, Pair-wise RL notably boosts performance on the Arena-Hard V2 benchmark, with an improvement from 66.6 to 73.8, showcasing its effectiveness in refining alignment. These results confirm that Pairwise RL sharpens preference boundaries, providing the informative supervision signals needed for overall improvement across various benchmarks.

\begin{table}[H]
\centering
\small
\caption{Improvements via General RL Training (SFT $\rightarrow$ Point-wise RL $\rightarrow$ Pair-wise RL)}
\label{tab:phase2_rl}
\resizebox{0.8 \textwidth}{!}{
\begin{tabular}{llccccc}
\toprule
{Domain} & {Benchmark} & {SFT} & {Point-wise RL} & {$\Delta$} & {Pair-wise RL} & {$\Delta$} \\ 
\midrule
{Code} & LCB V6 & 62.0 & \textbf{66.0} & +4.0 & 65.6 & -0.4 \\
\midrule
\multirow{2}{*}{{Alignment}} & Arena-Hard V2 & 60.2 & 66.6 & +6.4 & \textbf{73.8} & +7.2 \\
 & Multi-Challenge & 44.4 & 47.7 & +3.3 & \textbf{55.1} & +7.4 \\ 
\bottomrule
\end{tabular}
}
\end{table}

\subsection{Deep Search Ability}

In this section, we primarily focus on enhancing the \textbf{deep search capabilities} of our model, specifically on the data pipeline and the staged training process.
Deep search is defined as a retrieval-centric task that operates under complex multi-hop reasoning and extremely long-context settings. 
In this paradigm, models iteratively interact with the environment to acquire information, enabling them to solve challenging search problems.

\subsubsection{Data Construction}
\begin{figure}[!h] 
    \centering
    \includegraphics[width=1\textwidth]{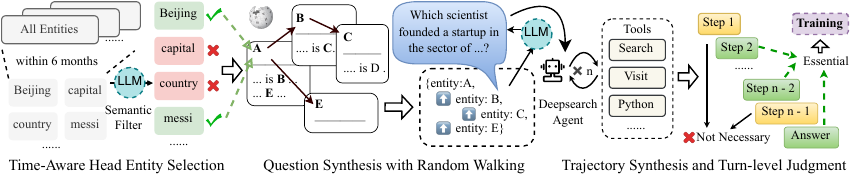} 
    \caption{A data construction pipeline for deep search, including complex multi-hop QA sampling and the synthesis of long-horizon reasoning trajectories.} 
    \label{fig:toolmindweb} 
\end{figure}

To enhance the search capability of our base model, we construct a large-scale, complex search dataset, which includes a substantial number of multi-hop QA pairs derived from entity-relation graphs built upon Wikipedia, as well as high-quality long-range search trajectories that have undergone multi-stage rigorous filtering.
The entire data construction pipeline is illustrated in Figure~\ref{fig:toolmindweb}. 
To facilitate further research, we have open-sourced the constructed dataset on HuggingFace~\footnote{https://huggingface.co/datasets/Nanbeige/ToolMind-Web-QA}.

\label{sec:posttraining}

\paratitle{Temporal-Aware Head Entity Selection and Question Synthesis with Random Walking.}
To ensure the timeliness and complexity of the synthesized QA data, we first extract informative head entities from Wikipedia that have been updated within the past six months.
Following the framework of \cite{yang2025toolmind}, we construct an entity-relation graph and perform conditional random walks to extract relational paths of a predefined length.
These chains, along with their detailed temporal contexts, are then fed into a robust LLM to synthesize intricate questions.

\paratitle{Trajectory Synthesis and Turn-level Judgment.}
To synthesize search trajectories, we employ multiple agent frameworks to address the generated queries, sampling a diverse set of reasoning paths.
These trajectories are subsequently mapped into multi-turn tool-invocation sequences from a unified agent perspective. To further guarantee the quality of the synthesized data, we implement a rigorous turn-level judgment mechanism. Specifically, we employ a critic model to evaluate each step of the interaction based on three dimensions: logical soundness, tool-call accuracy, and informational gain. 
Any turn that fails to meet these criteria does not participate in model training or provide a negative reward for the model. This fine-grained filtering ensures that the final trajectories provide a high-fidelity signal for supervised fine-tuning and preference alignment.

\subsubsection{Preliminary Performance}
\label{sec:search_performance}
To empirically validate the effectiveness of our proposed data construction pipeline, we conduct a controlled experiment using Nanbeige4-3B-2511 as the base model. 
Specifically, we train the model exclusively on the synthetic multi-hop QA and search trajectories generated via the methods described in Section~\ref{sec:posttraining}, intentionally excluding other open-source data.
We evaluate our model on a range of long-horizon benchmarks, including GAIA~\cite{mialon2023gaiabenchmarkgeneralai}, BrowseComp~\cite{wei2025browsecomp}, BrowseComp-ZH~\cite{zhou2025browsecomp}, Humanity's Last Exam~(HLE)~\cite{phan2025humanitysexam}, SEAL-0~\cite{pham2025sealqaraisingbarreasoning}, xBench-DeepSearch-2505~\cite{chen2025xbenchtrackingagentsproductivity}, and xBench-DeepSearch-2510~\cite{chen2025xbenchtrackingagentsproductivity}.
For HLE and GAIA, we only test and report the results of the text-only subset. 
In addition, HuggingFace has been explicitly disabled in these tools.

The model is evaluated within the Mindflow framework~\footnote{https://github.com/MiroMindAI/MiroThinker}, employing a suite of tools: Serper~\footnote{https://serper.dev/} for environment search, Jina~\footnote{https://jina.ai/} for webpage content extraction, and E2B Sandbox~\footnote{https://e2b.dev/} as the secure sandbox environment. A comparative analysis of our results across specific stages is presented below:

\begin{table}[!t]
\centering
\small
\caption{Preliminary evaluation results on search benchmarks for the synthetic QA.}
\label{tab:search}
\resizebox{1 \textwidth}{!}{
\begin{tabular}{lccccccc}
\toprule
Benchmark
& \makecell[c]{GAIA\\(text-only)}
& \makecell[c]{Browse\\Comp}
& \makecell[c]{Browse\\Comp-ZH}
& \makecell[c]{HLE\\(text-only)}
& \makecell[c]{SEAL-0}
& \makecell[c]{xBench\\DeepSearch-05}
& \makecell[c]{xBench\\DeepSearch-10}
\\
\midrule
Nanbeige4-3B-2511                  & 19.4  &    0.8   &    3.1   &  13.9 &  12.6    &     33.0       &          11.0    \\
\midrule
\quad + Synthetic QA & \textbf{58.3} & \textbf{14.4}      & \textbf{30.1}         & \textbf{22.4} & \textbf{36.0}   & \textbf{76.0}              & \textbf{30.0}       \\
\bottomrule
\end{tabular}
}
\end{table}

The quantitative results are summarized in Table~\ref{tab:search}. 
The incorporation of our synthetic data yields a significant performance improvement across all benchmarks compared to the base model. Notably, the model achieves a substantial leap on xBench-DeepSearch-2505 (improving from 33.0 to 76.0), demonstrating that our data synthesis pipeline effectively endows the model with robust multi-hop reasoning and long-context search abilities.

\subsection{Coding Ability}

In this section, we primarily focus on enhancing the \textbf{coding capabilities} of our base model, including the data construction pipeline, staged training strategy, and evaluation settings.

\subsubsection{Judge System}
We build a unified \textbf{judge system} that is shared across SFT data construction, RL data construction, and subsequent RL training and evaluation.
It combines a \textbf{multi-language sandbox} for execution-based correctness checking with a dedicated \textbf{instruct judge model} for time-complexity comparison.
The instruct model is specifically trained for fast complexity assessment in RL settings, enabling efficient online comparison between the predicted complexity of model-generated solutions and the reference optimal bound.

\subsubsection{Data Construction}
Our data construction consists of two components: SFT data construction with our judge system to filter time-optimal solutions, and RL data construction with on-policy difficulty filtering to improve sample efficiency.

\paragraph{SFT Data Construction.}
Our SFT data construction uses this judge system to assess solution quality from two key aspects: (1) \emph{functional correctness} by executing the program in a sandbox, and (2) \emph{time complexity} by combining execution signals with model-based complexity analysis.
During data generation, we sample multiple candidate solutions per problem.
The candidates are then verified by the judge system, and we keep those that are judged to be time-optimal (or among the best complexity class) for the given problem.

\paragraph{RL Data Construction.}
Each RL sample contains a problem statement, test cases, a time-complexity-optimal solution, and the corresponding optimal complexity label.
The optimal solution and complexity are obtained by prompting multiple strong LLMs and employing a strong LLM to synthesize the candidates into a single best solution, which are then used as supervision signals for reward shaping and difficulty control.

In both stages, we perform \textbf{on-policy filtering} by running multiple rollouts per problem ($n=8$) and selecting samples based on how many rollouts meet a stage-specific criterion.
In Stage~1, we use a \emph{difficulty-based} criterion: a problem is retained if the policy can solve it in a moderate number of rollouts (we keep problems with k in [1, 5] successful solves out of 8).
In Stage~2, we use a \emph{complexity-based} criterion: we count how many rollouts produce solutions whose estimated time complexity satisfies the target bound, and retain problems with k in [1,5] complexity-satisfying rollouts out of 8.

\subsubsection{Staged Training Process}
\begin{figure}[!t]
    \centering
    \includegraphics[width=0.9\textwidth]{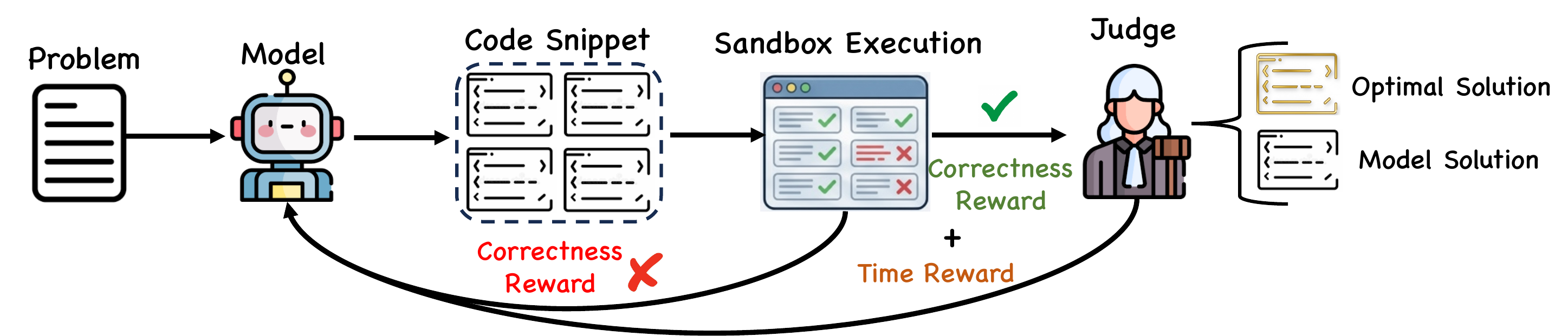}
    \caption{Gated time-complexity reward design in code RL. The time reward $R_{\mathrm{time}}$ is activated only when a solution passes all test cases ($\mathrm{PassRate}=1$), and the judge system provides online feedback by comparing the predicted time complexity against the reference optimal bound.}
    \label{fig:time_reward}
\end{figure}

Starting from the General-RL checkpoint, we further conduct two stages of code RL.
In Stage~1, we optimize for solution correctness using a pass-rate reward, defined as the fraction of test cases passed for each problem.
In Stage~2, after the policy can reliably solve problems, we additionally encourage higher-quality solutions by introducing a time-complexity reward \emph{only when all test cases are passed}; otherwise the reward reduces to correctness-only signals.
Specifically, the judge system provides online feedback by comparing the model's output against the reference optimal complexity and checking whether the generated solution matches the reference optimal solution when applicable.
As illustrated in Fig.~\ref{fig:time_reward}, the time-complexity reward is activated only for fully correct solutions.

\[
R=
\begin{cases}
R_{\mathrm{format}}+R_{\mathrm{correctness}}, & \mathrm{PassRate}<1,\\
R_{\mathrm{format}}+R_{\mathrm{correctness}}+R_{\mathrm{time}}, & \mathrm{PassRate}=1.
\end{cases}
\]

As a concrete illustration, Appendix~\ref{app:time_complexity_case_study} presents LiveCodeBench case studies comparing model outputs \textbf{before} vs.\ \textbf{after} the time-reward stage, highlighting typical complexity-class improvements.

\subsubsection{Training Dynamics}
\begin{figure}[!t]
    \centering
    \includegraphics[width=0.75\textwidth]{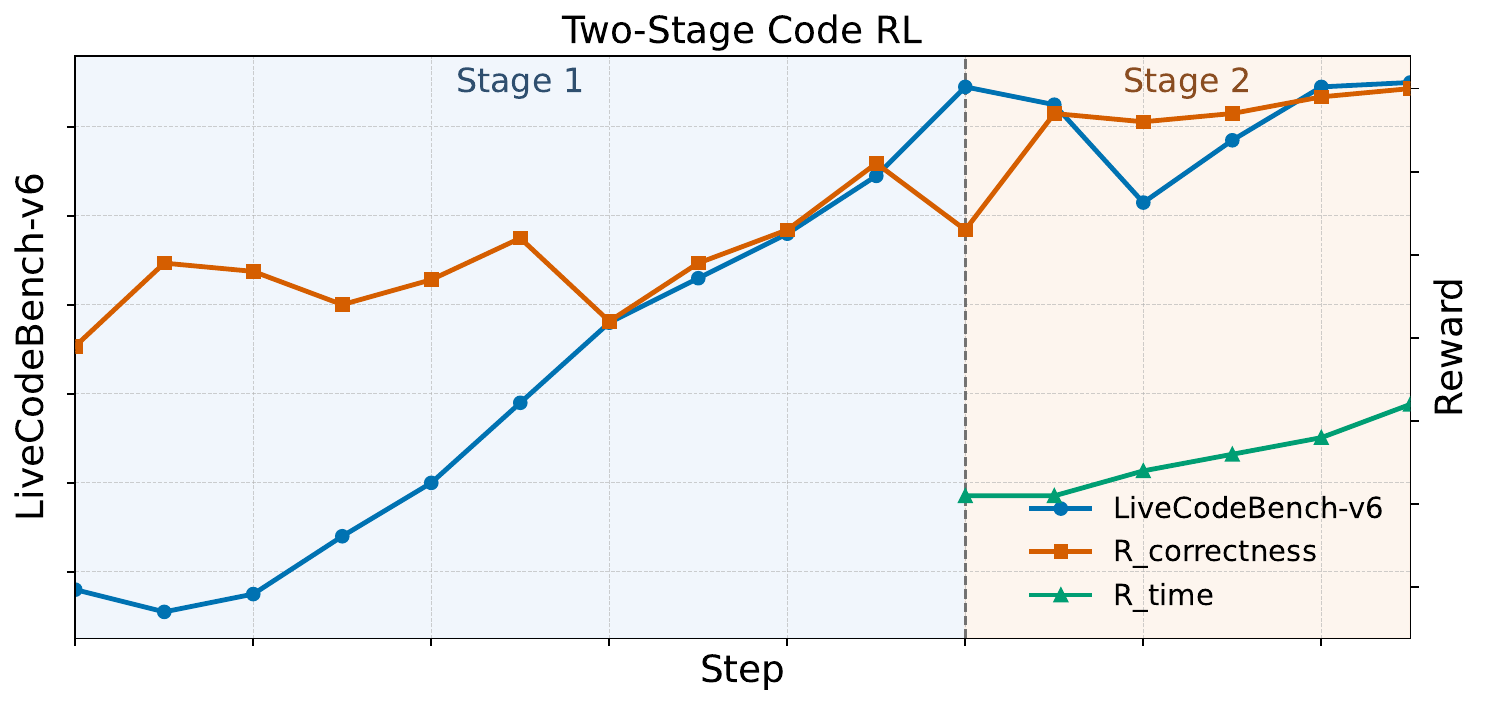}
    \caption{Training dynamics of two-stage code RL. We track the reward (including the gated $R_{\mathrm{time}}$ in Stage~2) and LiveCodeBench performance across training, showing consistent improvements from Stage~1 to Stage~2.}
    \label{fig:two_stage_code_rl}
\end{figure}

Across the two-stage code RL, we observe consistent improvements in both reward signals and downstream coding performance.
In Stage~1, the \emph{correctness reward} ($R_{\mathrm{correctness}}$) increases sharply, reflecting rapid gains in producing valid and correct solutions.
In Stage~2, $R_{\mathrm{correctness}}$ improves more modestly, while the gated \emph{time reward} ($R_{\mathrm{time}}$) rises substantially, indicating that the policy is indeed optimizing time complexity once correctness is largely achieved.
The overall reward and performance trends across the two-stage training are shown in Fig.~\ref{fig:two_stage_code_rl}.

\subsection{Whole Training Recipe of Nanbeige4.1}

Nanbeige4.1-3B is initialized from Nanbeige4-3B-Base and further optimized through a structured post-training pipeline combining large-scale SFT and cascaded RL.

We first conduct extended SFT, increasing the maximum context length from 64K to 256K compared to the previous Nanbeige4-3B-2511 version. This longer context window is essential for supporting long-horizon reasoning and multi-turn deep-search planning.

In the RL phase, we adopt a staged optimization strategy. General RL is performed sequentially with point-wise RL followed by pair-wise RL to enhance both standalone response quality and comparative preference alignment. Code RL is then conducted in two stages: a correctness stage that maximizes execution pass rate, followed by a gated time-complexity stage that activates efficiency rewards only when full correctness is achieved. Finally, we apply a lightweight agentic RL stage to strengthen tool-use and search behavior.

This unified training recipe enables Nanbeige4.1-3B to maintain strong domain-specific performance while emerging as a well-balanced generalist model under strict capacity constraints.

\section{Experiment}

Our evaluation comprises three components: 
general reasoning, deep-search agentic tasks, and a real-world coding challenge. General reasoning benchmarks measure core capability boundaries. Deep-search tasks evaluate long-horizon planning and tool-augmented multi-step reasoning in realistic environments. The real-world algorithm challenge provides an out-of-distribution stress test.

\subsection{General Task Evaluations}

For general reasoning capabilities, we evaluate across five major categories:
\textbf{code, math, science, alignment, and tool-use}.

\begin{itemize}[leftmargin=10pt]
    \item \textbf{Code.}
    We report results on LiveCodeBench-V5, LiveCodeBench-V6 \cite{jain2024livecodebenchholisticcontaminationfree}, and 
LiveCodeBench-Pro \cite{zheng2025livecodebench} to assess code generation ability and 
    execution-based correctness under increasing difficulty.

    \item \textbf{Mathematics.}
    We include IMO-Answer-Bench \cite{luong2025towards}, HMMT \cite{balunovic2025matharena}, and AIME-2026-I \footnote{https://huggingface.co/datasets/MathArena/aime\_2026\_I}
    to evaluate symbolic reasoning and competition-level problem solving.

    \item \textbf{Science.}
    We benchmark on GPQA \cite{rein2023gpqa} and HLE \cite{phan2025humanitysexam} to measure multi-step 
    scientific reasoning and domain knowledge integration.

    \item \textbf{Alignment.}
    We use Arena-Hard-V2 \cite{li2024arenahard} and Multi-Challenge \cite{deshpande2025multichallenge} to assess 
    preference modeling robustness and response quality under adversarial 
    or challenging prompts.

    \item \textbf{Tool-use.}
    We evaluate BFCL \cite{patil2025bfcl} and Tau2-Bench \cite{barres2025tau2bench}, which test 
    function-calling reliability and multi-step tool use capability.
\end{itemize}

For compared baseline models, we include open-source models at similar scale (Qwen3-4B-2507) and our previous release (Nanbeige4-3B-2511), 
as well as substantially larger open models including Qwen3-30B-A3B-2507, Qwen3-32B, and Qwen3-Next-80B-A3B. 
This setup allows us to assess both same-scale competitiveness and cross-scale efficiency.

\begin{table*}[t]
\centering
\small
\setlength{\tabcolsep}{3pt}
\caption{Evaluation results across code, math, science, alignment, and tool-use benchmarks.}
\resizebox{\textwidth}{!}{
\begin{tabular}{lcccccc}
\toprule
{Benchmark} 
& \makecell[c]{Qwen3-4B\\2507}
& \makecell[c]{Qwen3-32B}
& \makecell[c]{Qwen3-30B\\A3B-2507}
& \makecell[c]{Qwen3-Next\\-80B-A3B} 
& \makecell[c]{Nanbeige4-3B\\2511} 
& \makecell[c]{Nanbeige4.1-3B} \\
\midrule

\multicolumn{7}{c}{\textit{Code}} \\
\midrule

LCB-V6            & 57.4 & 55.7 & 66.0 & \underline{68.7} & 46.0 & \textbf{76.9} \\
LCB-Pro-Easy      & 40.2 & 42.3 & 60.8 & \underline{78.8} & 40.2 & \textbf{81.4} \\
LCB-Pro-Medium    & 5.3  & 3.5  & 3.5  & \underline{14.3} & 5.3  & \textbf{28.1} \\
\midrule

\multicolumn{7}{c}{\textit{Math}} \\
\midrule

AIME 2026 I       & 81.46 & 75.83 & 87.30 & \textbf{89.24} & 84.10 & \underline{87.40} \\
HMMT Nov          & 68.33 & 57.08 & 71.25 & \textbf{81.67} & 66.67 & \underline{77.92} \\
IMO-Answer-Bench  & 48.00 & 43.94 & \underline{54.34} & \textbf{58.00} & 38.25 & 53.38 \\
\midrule

\multicolumn{7}{c}{\textit{Science}} \\
\midrule

GPQA              & 65.8 & 68.4 & 73.4 & 77.2 & \underline{82.2} & \textbf{83.8} \\
HLE (Text-only)   & 6.72 & 9.31 & 11.77 & \textbf{13.70} & 10.98 & \underline{12.60} \\
\midrule

\multicolumn{7}{c}{\textit{Alignment}} \\
\midrule

Arena-Hard-V2     & 34.9 & 56.0 & 60.2 & \underline{62.3} & 60.0 & \textbf{73.2} \\
Multi-Challenge   & 41.14 & 38.72 & 49.40 & \textbf{56.52} & 41.20 & \underline{52.21} \\
\midrule

\multicolumn{7}{c}{\textit{Tool Use}} \\
\midrule

BFCL-V4           & 44.87 & 47.90 & 48.60 & 50.51 & \underline{53.80} & \textbf{56.50} \\
Tau2-Bench        & 45.90 & 45.26 & 47.70 & \textbf{57.40} & 41.77 & \underline{48.57} \\
\bottomrule
\end{tabular}
}
\label{tab:overall_benchmarks}
\end{table*}

\paragraph{Overall Results.}

Table~\ref{tab:overall_benchmarks} shows that Nanbeige4.1-3B 
substantially outperforms both Qwen3-4B-2507 and its predecessor 
Nanbeige4-3B-2511 across all evaluated domains, 
demonstrating the effectiveness of our post-training strategy.

More notably, despite having only 3B parameters, 
Nanbeige4.1-3B consistently surpasses 30B--32B class models 
(Qwen3-30B-A3B-2507 and Qwen3-32B) on the majority of benchmarks, 
including coding, alignment, and tool-use tasks. 
The gains are particularly pronounced on execution-based coding benchmarks 
such as LiveCodeBench-V6 and LiveCodeBench-Pro-Medium, 
where Nanbeige4.1-3B achieves large absolute margins. When compared to Qwen3-Next-80B-A3B, 
Nanbeige4.1-3B remains competitive and exhibits complementary strengths: 
it leads on several coding and alignment benchmarks, 
while the 80B model retains advantages on selected mathematics and tool-use tasks. 

These results suggest that strong cross-domain reasoning performance 
can be achieved through targeted post-training and agent-oriented optimization, 
even at a significantly smaller parameter scale.

\subsection{Deep Search Task Evaluations}

In this section, we present a comprehensive evaluation of the performance of Nanbeige4.1-3B on deep search tasks. We evaluate its capabilities in handling complex tasks by benchmarking against general foundation models of comparable size, specialized search-agent models, and significantly larger-scale foundation models.

\begin{table}[t]
\centering
\small
\caption{Performance comparison across various deep search benchmarks.}
\label{tab:nbg4-3b-evaluation}
\setlength{\tabcolsep}{4pt}
\resizebox{\textwidth}{!}{
\begin{tabular}{lccccccc}
\toprule
Benchmark
& \makecell[c]{GAIA\\(text-only)}
& \makecell[c]{Browse\\Comp}
& \makecell[c]{Browse\\Comp-ZH}
& \makecell[c]{HLE\\(text-only)}
& \makecell[c]{SEAL-0}
& \makecell[c]{xBench\\DeepSearch-05}
& \makecell[c]{xBench\\DeepSearch-10}
\\
\midrule
\multicolumn{8}{c}{Research Agents} \\
\midrule
Tongyi-DeepResearch-30B & 70.90 & 43.40 & 46.70 & 32.90 & -     & 75.00 & -     \\
MiroThinker-v1.0-8B     & 66.40 & 31.10 & 40.20 & 21.50 & 40.40 & 60.60 & -     \\
AgentCPM-Explore-4B     & 63.90 & 25.00 & 29.00 & 19.10 & 40.00 & 70.00 & -     \\
\midrule
\multicolumn{8}{c}{Large Foundation Models with Tools} \\
\midrule
GLM-4.6-357B       & 71.90 & 45.10 & 49.50 & 30.40 & -     & 70.00 & -     \\
Minimax-M2-230B    & 75.70 & 44.00 & 48.50 & 31.80 & -     & 72.00 & -     \\
DeepSeek-V3.2-671B & 63.50 & 67.60 & 65.00 & 40.80 & 38.50 & 71.00 & -     \\
\midrule
\multicolumn{8}{c}{Small Foundation Models with Tools} \\
\midrule
Qwen3-4B-2507         & 28.33 & 1.57  & 7.92  & 11.13 & 15.74 & \underline{34.00} & 5.00  \\
Qwen3-8B              & 19.53 & 0.79  & 5.15  & 10.24 & 6.34  & 31.00 & 2.00  \\
Qwen3-14B             & 30.23 & 2.36  & 7.11  & 10.17 & 12.64 & 34.00 & 9.00  \\
Qwen3-32B             & 30.17 & 3.15  & 7.34  & 9.26  & 8.15  & 39.00 & 8.00  \\
Qwen3-30B-A3B-2507    & 31.63 & 1.57  & 4.12  & \underline{14.81} & 9.24 & 25.00 & 10.00 \\
Qwen3-Next-80B-A3B    & \underline{34.02} & \underline{5.60} & \underline{8.25} & 9.26 & \underline{18.18} & 27.00 & 6.00 \\
\midrule
\multicolumn{8}{c}{Ours} \\
\midrule
Nanbeige4-3B-2511 & 19.42 & 0.79  & 3.09  & 13.89 & 12.61 & 33.00 & \underline{11.00} \\
Nanbeige4.1-3B    & \textbf{69.90} & \textbf{19.12} & \textbf{31.83} & \textbf{22.29} & \textbf{41.44} & \textbf{75.00} & \textbf{39.00} \\
\bottomrule
\end{tabular}
}
\end{table}

\subsubsection{Experimental Setup and Comparison Baselines}
All experimental settings are the same as those of Section~\ref{sec:search_performance}.
These benchmarks require iterative retrieval, planning, tool interaction, and cross-document reasoning, reflecting realistic autonomous agent workloads.
To thoroughly contextualize the performance of Nanbeige4.1-3B, we conduct a diverse comparison against several categories of existing models. This includes general-purpose foundation models equipped with tools (e.g., Qwen-series models), specialized search-agent models such as MiroThinker-8B, AgentCPM-Explore-4B, and Tongyi-DeepResearch-30B. We also incorporate large-scale open foundation models exceeding 100B parameters, comprising Minimax-M2.1, GLM-4.6, and DeepSeek-V3.2 for comparison. 

\subsubsection{Overall Performance Analysis}

Table \ref{tab:nbg4-3b-evaluation} presents a comprehensive performance comparison across various search agent benchmarks. While "Research Agents" (e.g., Tongyi-DeepResearch-30B) and "Large Foundation Models with Tools" (e.g., Minimax-M2, DeepSeek-V3.2) generally exhibit strong capabilities in their respective categories, particularly on GAIA and browsing tasks, the "Small Foundation Models with Tools" (Qwen3 series) typically show lower performance levels.

Our model, Nanbeige4.1-3B, demonstrates a remarkable leap in performance compared to its baseline Nanbeige4-3B-2511 and significantly outperforms other small foundation models with tools across all benchmarks. More critically, Nanbeige4.1-3B achieves state-of-the-art results across nearly all evaluated benchmarks, including GAIA (69.90), xBench-DeepSearch-05 (75.00), and SEAL-0 (41.44). These scores not only surpass its direct competitors in the small model category but also place it on par with, or even exceeding, the performance of many larger "Research Agents" and "Large Foundation Models with Tools." This robust performance underscores Nanbeige4.1-3B's exceptional effectiveness in complex search and agentic tasks, demonstrating that competitive capabilities can be achieved efficiently even with a smaller model size through our proposed methodology.

\subsection{Real-world Algorithmic Challenges: LeetCode Weekly Contests}
\label{sec:leetcode_contests}

Beyond curated academic benchmarks, we further evaluate Nanbeige4.1-3B on real-world competitive programming tasks drawn from recent LeetCode weekly contests~\footnote{https://leetcode.cn/contest/}, providing a practical out-of-distribution stress test of algorithmic reasoning ability.

We apply Nanbeige4.1-3B, alongside Qwen models of varying scales, to solve contest problems under standard competitive programming settings. The generated solutions are directly submitted to the official LeetCode platform, and performance is measured by the final acceptance rate.

\begin{table}[ht]
\centering
\small
\setlength{\tabcolsep}{5pt} 
\caption{Comparison of Pass Rates on LeetCode Weekly Contests 484--488.}
\label{tab:leetcode_results}
\begin{tabular}{lcccc}
\toprule
{Benchmark} & 
\makecell[c]{Qwen3-4B-2507} & 
\makecell[c]{Qwen3-32B} & 
\makecell[c]{Qwen3-30B-A3B-2507} & 
\makecell[c]{Nanbeige4.1-3B} \\
\midrule
LeetCode Pass Rate (\%) & 55.0 & 50.0 & \underline{65.0} & \textbf{85.0} \\
\bottomrule
\end{tabular}
\end{table}

As shown in Table~\ref{tab:leetcode_results}, Nanbeige4.1-3B successfully solves 17 out of 20 problems, achieving an overall pass rate of 85.0\%. In virtual participation mode, where ranking is determined by correctness and time efficiency, our model achieves 1st place in Weekly Contest 487 and 3rd place in Weekly Contest 488. Compared to Qwen3 models of similar or substantially larger scales, Nanbeige4.1-3B demonstrates a clear performance advantage on these real-world contest tasks. Detailed problem statements and representative solution examples are provided in Appendix~\ref{app:leetcode_examples}.

\section{Conclusion}

In this work, we present Nanbeige4.1-3B, a unified generalist model that demonstrates advanced capabilities in reasoning, coding, and long-horizon search. We achieve this by integrating point-wise and pair-wise reward modeling for precise preference alignment, optimizing code generation for both correctness and computational efficiency, and incorporating both turn and trajectory level signals to enable robust long-horizon agentic planning. Consequently, Nanbeige4.1-3B consistently outperforms comparable small models and remains competitive with larger baselines across diverse benchmarks and real-world challenges.

Looking ahead, we aim to push the boundaries of compact models in complex coding and research agent scenarios. Concurrently, we focus on improving inference efficiency by enabling tasks to be solved with shorter outputs and fewer tool invocations, while exploring architectural innovations to further enhance the potential of the compact generalist model.

\newpage

\bibliographystyle{plain} 
\bibliography{references}
\newpage
\appendix
\section{Author List}

Authors are listed in \textbf{alphabetical order by first name}. Names marked with an asterisk (*) denote individuals who were previously affiliated with our team. Yang Song is the corresponding author and can be reached at \texttt{songyang@kanzhun.com}.

Chen Yang, Guangyue Peng, Jiaying Zhu, Ran Le, Ruixiang Feng, Tao Zhang, Xiyun Xu, Yang Song\textsuperscript{\dag}, Yiming Jia, Yuntao Wen, Yunzhi Xu, Zekai Wang, Zhenwei An\textsuperscript{*}, Zhicong Sun, Zongchao Chen

\section{Leetcode Contest Examples}
\label{app:leetcode_examples}

The following examples are drawn from LeetCode Weekly Contest 488 under virtual participation. Since the contest was conducted on the Chinese version of LeetCode, the original problem statements presented to the model were in Chinese, and the generated code contained Chinese comments. For clarity, we provide English translations of both the problem descriptions and the model's comments below.

\subsection{Problem Q1: Count Dominant Indices}

\begin{tcolorbox}[colback=gray!5, colframe=gray!60, title=\textbf{Problem Description}]
\textbf{Given:} An integer array \texttt{nums} of length $n$.

\textbf{Definition:} An index $i$ is called a \textit{dominant index} if the following condition is met:
\[
\text{nums}[i] > \text{average}(\text{nums}[i + 1], \text{nums}[i + 2], \dots, \text{nums}[n - 1])
\]
Your task is to count the number of \textit{dominant indices}.

\textbf{Note:}
\begin{itemize}
    \item The average is the sum of the group divided by the number of elements.
    \item The rightmost element of the array is not considered a dominant index.
\end{itemize}
\end{tcolorbox}

\vspace{0.5cm}
\noindent\textbf{Model Output (Nanbeige4.1-3B):}

The model correctly transformed the inequality to avoid floating-point precision issues and utilized a suffix sum approach to solve the problem in $O(n)$ time complexity.

\begin{lstlisting}{python}
from typing import List

class Solution:
    def dominantIndices(self, nums: List[int]) -> int:
        """
        [Translated Comment] Count the number of dominant indices in the array.
        Dominant Index Definition: Index i such that:
        nums[i] > average(nums[i+1], ..., nums[n-1])
        Note: The rightmost element (index n-1) is not considered a candidate.
        
        Optimization Strategy:
        1. Avoid floating-point arithmetic: Transform nums[i] > sum / len 
           into nums[i] * len > sum.
        2. Traverse from right to left, dynamically maintaining the suffix sum.
        3. Time Complexity: O(n), Space Complexity: O(1).
        """
        n = len(nums)
        # Boundary condition: No valid dominant index if length <= 1
        if n <= 1:
            return 0
        
        count = 0
        suffix_sum = 0  # Stores sum of nums[i+1] to nums[n-1]
        
        # Traverse backwards from the second to last element to index 0
        for i in range(n - 2, -1, -1):
            suffix_sum += nums[i + 1]  # Accumulate current right-side part
            length = n - i - 1         # Number of elements on the right
            
            # Strict inequality check (avoiding floating point issues)
            if nums[i] * length > suffix_sum:
                count += 1
        
        return count
\end{minted}
\end{lstlisting}

\subsection{Problem Q2: Apply Operations to Maximize Array Sum}

\begin{tcolorbox}[colback=gray!5, colframe=gray!60, title=\textbf{Problem Description}]
\textbf{Given:} An integer array \texttt{nums}.

\textbf{Operation:} Repeatedly perform the following until no more operations can be done:
\begin{itemize}
    \item If there are two adjacent and equal elements in the array, select the \textbf{leftmost} such pair.
    \item Replace them with their \textbf{sum}.
\end{itemize}
The array size decreases by 1 after each merge.

\textbf{Task:} Return the final array after all possible merge operations are completed.
\end{tcolorbox}

\vspace{0.5cm}
\noindent\textbf{Model Output (Nanbeige4.1-3B):}

The model employs a stack-based simulation to handle the "leftmost" merge constraint efficiently. This mimics the recursive merge process in a single pass with $O(n)$ time complexity.

\begin{lstlisting}{python}
from typing import List

class Solution:
    def mergeAdjacent(self, nums: List[int]) -> List[int]:
        """
        [Translated Comment] Simulate the merge process using a stack:
        - Traverse the array, comparing each element with the stack top.
        - If equal, continue merging (pop stack top and accumulate) until 
          not equal or stack is empty.
        - Push the merge result (or original element) onto the stack.
        - The final stack content is the result after no more merges can be done.
        
        Time Complexity: O(n), each element is pushed/popped at most once.
        Space Complexity: O(n), max stack depth.
        """
        stack = []
        for num in nums:
            current = num
            # Continuous merge: when stack top equals current, pop and accumulate
            while stack and stack[-1] == current:
                current += stack.pop()
            stack.append(current)
        return stack
\end{minted}

\subsection{Problem Q3: Count Subarrays with Cost Constraint}

\begin{tcolorbox}[colback=gray!5, colframe=gray!60, title=\textbf{Problem Description}]
\textbf{Given:} An integer array \texttt{nums} and an integer $k$.

\textbf{Definition:} The \textit{cost} of a subarray $\text{nums}[l..r]$ is defined as:
\[
\text{cost} = (\max(\text{nums}[l..r]) - \min(\text{nums}[l..r])) \times (r - l + 1)
\]
\textbf{Task:} Return the number of subarrays where the cost is less than or equal to $k$.
\end{lstlisting}

\vspace{0.5cm}
\noindent\textbf{Model Output (Nanbeige4.1-3B):}

The model correctly identifies this as a sliding window problem. To maintain the maximum and minimum values within the moving window efficiently, it utilizes two monotonic deques, achieving an optimal solution.

\begin{lstlisting}{python}
from collections import deque
from typing import List

class Solution:
    def countSubarrays(self, nums: List[int], k: int) -> int:
        # [Translated] Boundary check: empty array or negative k (cost is non-negative)
        if not nums or k < 0:
            return 0
        
        n = len(nums)
        dq_max = deque()  # Monotonic decreasing deque: stores indices of max values
        dq_min = deque()  # Monotonic increasing deque: stores indices of min values
        l = 0
        ans = 0
        
        for r in range(n):
            # Maintain max deque (pop back if <= current value)
            while dq_max and nums[dq_max[-1]] <= nums[r]:
                dq_max.pop()
            dq_max.append(r)
            
            # Maintain min deque (pop back if >= current value)
            while dq_min and nums[dq_min[-1]] >= nums[r]:
                dq_min.pop()
            dq_min.append(r)
            
            # Calculate cost for current window [l, r]
            cur_max = nums[dq_max[0]]
            cur_min = nums[dq_min[0]]
            length = r - l + 1
            cost = (cur_max - cur_min) * length
            
            # Shrink left boundary: if cost > k, move l to the right
            while cost > k and l <= r:
                # Remove left boundary influence on extremums
                if dq_max[0] == l:
                    dq_max.popleft()
                if dq_min[0] == l:
                    dq_min.popleft()
                l += 1
                if l > r:
                    break
                # Recalculate cost after shrinking
                cur_max = nums[dq_max[0]]
                cur_min = nums[dq_min[0]]
                length = r - l + 1
                cost = (cur_max - cur_min) * length
            
            # Accumulate all valid subarrays ending at r
            if l <= r:
                ans += (r - l + 1)
        
        return ans
\end{lstlisting}

\subsection{Problem Q4: Maximize Score of K Pairs}

\begin{tcolorbox}[colback=gray!5, colframe=gray!60, title=\textbf{Problem Description}]
\textbf{Given:} Two integer arrays \texttt{nums1} (length $n$) and \texttt{nums2} (length $m$), and an integer $k$.

\textbf{Task:} Choose exactly $k$ pairs of indices $(i_1, j_1), (i_2, j_2), \dots, (i_k, j_k)$ such that:
\[
0 \le i_1 < i_2 < \dots < i_k < n \quad \text{and} \quad 0 \le j_1 < j_2 < \dots < j_k < m
\]
The score for each pair $(i, j)$ is $\text{nums1}[i] \times \text{nums2}[j]$.
Return the \textbf{maximum total score} (sum of products) achievable.
\end{tcolorbox}

\vspace{0.5cm}
\noindent\textbf{Model Output (Nanbeige4.1-3B):}

The model proposes a Dynamic Programming approach using rolling arrays to optimize space. It constructs a 2D state transition `curr[i][j]` representing the max score selecting a specific number of pairs from the prefixes of both arrays.

\begin{lstlisting}{python}
from typing import List

class Solution:
    def maxScore(self, nums1: List[int], nums2: List[int], k: int) -> int:
        n, m = len(nums1), len(nums2)
        # [Translated] Edge case: k=0 returns 0
        if k == 0:
            return 0
        # Implicit constraint check n >= k and m >= k
        if k > n or k > m:
            return 0
        
        # prev[i][j]: Max score choosing t-1 pairs from first i and first j elements
        prev = [[0] * (m + 1) for _ in range(n + 1)]
        NEG_INF = float('-inf')
        
        # Iterate to select 1 to k pairs
        for _ in range(1, k + 1):
            # curr initialized to -inf
            curr = [[NEG_INF] * (m + 1) for _ in range(n + 1)]
            for i in range(1, n + 1):
                num1 = nums1[i - 1]
                for j in range(1, m + 1):
                    num2 = nums2[j - 1]
                    # Three choices:
                    # 1. Skip current element of nums1: state from curr[i-1][j]
                    # 2. Skip current element of nums2: state from curr[i][j-1]
                    # 3. Take current pair (i-1, j-1): state from prev[i-1][j-1] + product
                    #    (ensures strictly increasing indices)
                    skip_i = curr[i - 1][j]
                    skip_j = curr[i][j - 1]
                    take = prev[i - 1][j - 1] + num1 * num2
                    curr[i][j] = max(skip_i, skip_j, take)
            prev = curr  # Rolling update: current layer becomes prev for next round
        
        # Final result: Max score choosing exactly k pairs from n and m elements
        return int(prev[n][m])
\end{lstlisting}

\section{Time-Complexity Case Study}
\label{app:time_complexity_case_study}

\paragraph{Scope and purpose.}
This section presents a small set of \textbf{time-complexity-focused case studies} comparing model outputs \textbf{before} vs.\ \textbf{after} introducing the \emph{gated time reward} in code RL. Concretely, we use examples from \textbf{LiveCodeBench} where the two models produce functionally correct solutions but differ in their algorithmic efficiency, and we analyze how the time-reward stage encourages outputs closer to the time-optimal complexity class.

\subsection{AtCoder ABC388 Excerpt: Simultaneous Kagamimochi}

\begin{tcolorbox}[colback=gray!5, colframe=gray!60, title=\textbf{Problem Description (Excerpt)}]
\textbf{Given:} $N$ mochi with non-decreasing sizes $A_1 \le A_2 \le \dots \le A_N$. \\
\textbf{Rule:} You can stack mochi of sizes $a$ (top) and $b$ (bottom) iff $a \le \frac{b}{2}$. \\
\textbf{Task:} Maximize the number $K$ of kagamimochi that can be made simultaneously by choosing $2K$ mochi and pairing them.
\end{tcolorbox}

\paragraph{Before (Candidate Code).}
\begin{lstlisting}{python}
import sys

def main():
    data = sys.stdin.read().strip().split()
    if not data:
        return
    N = int(data[0])
    A = list(map(int, data[1:1+N]))
    # The input is already sorted, but we sort to be safe.
    A.sort()
    low = 0
    high = N // 2
    while low < high:
        mid = (low + high + 1) // 2
        ok = True
        for i in range(mid):
            if 2 * A[i] > A[N - mid + i]:
                ok = False
                break
        if ok:
            low = mid
        else:
            high = mid - 1
    print(low)

if __name__ == "__main__":
    main()
\end{lstlisting}

\paragraph{After (Candidate Code).}
\begin{lstlisting}{python}
import sys

def main():
    data = sys.stdin.read().strip().split()
    if not data:
        return
    N = int(data[0])
    A = list(map(int, data[1:1+N]))
    # A is already sorted in non-decreasing order as per input specification.
    i = 0
    j = N // 2
    count = 0
    while i < N // 2 and j < N:
        if A[i] * 2 <= A[j]:
            count += 1
            i += 1
            j += 1
        else:
            j += 1
    print(count)

if __name__ == "__main__":
    main()
\end{lstlisting}

\subsubsection{Complexity Analysis}

\paragraph{Key observation.}
Because $A$ is sorted, the problem reduces to matching a \emph{small} mochi as the top with a \emph{large} mochi as the bottom, under the constraint $2\cdot A[\text{top}] \le A[\text{bottom}]$, without reusing mochi.

\paragraph{Before: binary search over the answer.}
The \texttt{before} solution binary-searches $K$ in $[0, \lfloor N/2 \rfloor]$. For each candidate $K=\texttt{mid}$, it checks feasibility by scanning $i=0..K-1$ and verifying $2A[i] \le A[N-K+i]$.
\begin{itemize}
    \item If we include the extra \texttt{sort()}, the time is $O(N\log N) + O\!\left(N \log N\right)=O(N\log N)$.
    \item If we use the input guarantee (already sorted) and drop \texttt{sort()}, the dominating part is the feasibility checks:
    \[
    O\!\left(\sum_{\text{iterations}} \texttt{mid}\right) = O\!\left(N \log N\right),
    \]
    because there are $O(\log N)$ iterations and each check costs $O(\texttt{mid}) \le O(N)$.
    \item Space complexity is $O(N)$ for storing $A$.
\end{itemize}

\paragraph{After: two-pointers greedy matching.}
The \texttt{after} solution uses two pointers: $i$ over the smaller half ($[0, \lfloor N/2 \rfloor)$) and $j$ over the larger half ($[\lfloor N/2 \rfloor, N)$). Pointer $j$ only moves forward, and $i$ moves forward exactly when a valid pair is formed.
\begin{itemize}
    \item Each pointer advances at most $N$ times, so the total time is $O(N)$ (no sorting is needed).
    \item Space complexity remains $O(N)$ for the array.
\end{itemize}

\paragraph{Conclusion.}
Both solutions are correct, but \texttt{after} improves the core algorithmic cost from $O(N\log N)$ to $O(N)$ by eliminating the binary search and performing a single-pass greedy matching enabled by the sorted input.

\subsection{AtCoder ARC195 A Excerpt: Twice Subsequence}
\begin{tcolorbox}[colback=gray!5, colframe=gray!60, title=\textbf{Problem Description (Excerpt)}]
\textbf{Given:} A sequence $A=(A_1,\dots,A_N)$ and a target sequence $B=(B_1,\dots,B_M)$ with $1\le M\le N\le 2\times 10^5$.\\
\textbf{Task:} Determine whether there exist \emph{at least two} subsequences of $A$ that match $B$.\\
Two subsequences are considered different if they use different sets of positions in $A$, even if the resulting values are identical.
\end{tcolorbox}

\paragraph{Before (Candidate Code).}
\begin{lstlisting}{python}
import sys
import bisect
from collections import defaultdict

def main():
    data = sys.stdin.read().strip().split()
    if not data:
        return
    it = iter(data)
    N = int(next(it))
    M = int(next(it))
    A = [int(next(it)) for _ in range(N)]
    B = [int(next(it)) for _ in range(M)]

    # Map each value to the sorted list of indices where it appears in A
    pos_map = defaultdict(list)
    for i, val in enumerate(A):
        pos_map[val].append(i)

    # Find the leftmost (earliest) matching subsequence
    left = []
    pos = -1
    for b in B:
        if b not in pos_map:
            print("No")
            return
        lst = pos_map[b]
        # Find first index > pos
        idx = bisect.bisect_left(lst, pos + 1)
        if idx == len(lst):
            print("No")
            return
        i = lst[idx]
        left.append(i)
        pos = i

    # Find the rightmost (latest) matching subsequence
    right = []
    pos = N
    for b in reversed(B):
        if b not in pos_map:
            print("No")
            return
        lst = pos_map[b]
        # Find last index < pos
        idx = bisect.bisect_left(lst, pos)
        if idx == 0:
            print("No")
            return
        i = lst[idx - 1]
        right.append(i)
        pos = i
    right.reverse()  # restore order to match B

    # If the two subsequences differ in at least one position, answer Yes
    for l, r in zip(left, right):
        if l != r:
            print("Yes")
            return
    print("No")

if __name__ == "__main__":
    main()
\end{lstlisting}

\paragraph{After (Candidate Code).}
\begin{lstlisting}{python}
import sys

def main():
    data = sys.stdin.read().split()
    if not data:
        return
    it = iter(data)
    N = int(next(it))
    M = int(next(it))
    A = [int(next(it)) for _ in range(N)]
    B = [int(next(it)) for _ in range(M)]
    
    # leftmost matching indices
    left = []
    j = 0
    for i in range(N):
        if j < M and A[i] == B[j]:
            left.append(i)
            j += 1
            if j == M:
                break
    if j < M:
        print("No")
        return
    
    # rightmost matching indices
    right = []
    j = M - 1
    for i in range(N-1, -1, -1):
        if j >= 0 and A[i] == B[j]:
            right.append(i)
            j -= 1
            if j < 0:
                break
    right.reverse()
    
    # compare left and right
    for k in range(M):
        if left[k] != right[k]:
            print("Yes")
            return
    print("No")

if __name__ == "__main__":
    main()
\end{lstlisting}

\subsubsection{Complexity Analysis}

\paragraph{Key idea: earliest vs. latest embedding.}
Let $L=(\ell_1,\dots,\ell_M)$ be the \emph{leftmost} embedding of $B$ in $A$, obtained greedily by scanning $A$ left-to-right and matching $B_1,B_2,\dots$ as early as possible.
Let $R=(r_1,\dots,r_M)$ be the \emph{rightmost} embedding of $B$ in $A$, obtained greedily by scanning $A$ right-to-left and matching $B_M,B_{M-1},\dots$ as late as possible.

If there exists at least one $k$ such that $\ell_k \ne r_k$, then we can construct two different matching subsequences:
one using the prefix choices from $L$ up to some point, and another using the suffix choices from $R$ (intuitively, there is ``slack'' at position $k$ allowing a different choice of index while still completing the match).
Conversely, if $L=R$ coordinate-wise, then every match is forced to use the same index for each $B_k$, so there is \emph{exactly one} matching subsequence.

\paragraph{Before: value-to-positions map + binary search.}
The \texttt{before} approach preprocesses a dictionary \texttt{pos\_map} mapping each value $v$ to the sorted list of indices where $A_i=v$.
Then:
\begin{itemize}
  \item It constructs $L$ by, for each $B_k$, binary searching the first occurrence index $> \ell_{k-1}$.
  \item It constructs $R$ similarly from the right, binary searching the last occurrence index $< r_{k+1}$.
  \item Finally it checks whether $L$ and $R$ differ in any coordinate.
\end{itemize}
\textbf{Time.} Building \texttt{pos\_map} costs $O(N)$. Each of the $M$ steps performs one binary search on an occurrence list, costing $O(\log N)$ in the worst case. Total time is $O(N + M\log N)$ per test case.\\
\textbf{Space.} The map stores all indices once, so $O(N)$.

\paragraph{After: two-pointer greedy scans.}
The \texttt{after} approach eliminates the map and binary searches:
\begin{itemize}
  \item Construct $L$ with a single left-to-right scan of $A$ using a pointer $j$ over $B$.
  \item Construct $R$ with a single right-to-left scan of $A$ using a pointer $j$ over $B$.
  \item Compare $L$ and $R$; output \texttt{Yes} if they differ, else \texttt{No}.
\end{itemize}
\textbf{Time.} Each scan is $O(N)$, and the final comparison is $O(M)$, hence total $O(N+M)=O(N)$ per test case.\\
\textbf{Space.} Arrays \texttt{left} and \texttt{right} store $M$ indices, so $O(M)$ (and thus $O(N)$).

\paragraph{Conclusion.}
Both solutions rely on the same correctness criterion: there are at least two matching subsequences iff the leftmost and rightmost embeddings differ at some position.
The \texttt{after} solution achieves optimal linear time by replacing the index-map + binary-search steps with two greedy linear passes, reducing the per-test-case complexity from $O(N + M\log N)$ to $O(N)$ while keeping $O(N)$ space.

\subsection{AtCoder ARC194 Excerpt: Minimum Cost Sort}

\begin{tcolorbox}[colback=gray!5, colframe=gray!60, title=\textbf{Problem Description (Excerpt)}]
\textbf{Given:} A permutation $P$ of $(1,2,\dots,N)$.\\
\textbf{Operation:} For $1 \le i \le N-1$, swap adjacent elements $(P_i,P_{i+1})$ with cost $i$.\\
\textbf{Task:} Compute the minimum total cost to sort $P$ in ascending order.
\end{tcolorbox}

\paragraph{Before (Candidate Code).}
\begin{lstlisting}{python}
import sys
import heapq

def solve():
    data = sys.stdin.read().strip().split()
    if not data:
        return
    it = iter(data)
    N = int(next(it))
    P = [0] + [int(next(it)) for _ in range(N)]  # 1-indexed

    # Initialize a min-heap with all adjacent inversions
    heap = []
    for i in range(1, N):
        if P[i] > P[i + 1]:
            heapq.heappush(heap, i)

    total_cost = 0
    while heap:
        i = heapq.heappop(heap)
        # Check if the inversion still exists (it might have been fixed already)
        if not (P[i] > P[i + 1]):
            continue
        # Perform the swap
        P[i], P[i + 1] = P[i + 1], P[i]
        total_cost += i
        # After swapping, check the three affected positions for new inversions
        for idx in (i - 1, i, i + 1):
            if 1 <= idx < N and P[idx] > P[idx + 1]:
                heapq.heappush(heap, idx)

    print(total_cost)

if __name__ == "__main__":
    solve()
\end{minted}

\paragraph{After (Candidate Code).}
\begin{minted}[linenos, frame=lines, framesep=2mm, fontsize=\small, breaklines]{python}
import sys

def solve() -> None:
    input = sys.stdin.readline
    N = int(input())
    P = list(map(int, input().split()))

    # Fenwick tree (1-indexed)
    bit = [0] * (N + 2)

    def update(i: int, delta: int) -> None:
        while i <= N:
            bit[i] += delta
            i += i & -i

    def query(i: int) -> int:
        s = 0
        while i > 0:
            s += bit[i]
            i -= i & -i
        return s

    total_cost = 0
    for val in P:
        # number of smaller elements to the left of 'val'
        cnt = query(val - 1)
        p0 = cnt + 1          # position of 'val' after larger elements to its left have passed
        if p0 < val:
            terms = val - p0   # number of right moves needed
            # sum from p0 to val-1 inclusive
            total_cost += (p0 + val - 1) * terms // 2
        update(val, 1)

    print(total_cost)

if __name__ == "__main__":
    solve()
\end{lstlisting}

\subsubsection{Complexity Analysis}

\paragraph{Before: heap-driven local bubble swaps.}
The \texttt{before} solution repeatedly fixes adjacent inversions, always picking the smallest index inversion via a min-heap, and then re-checks nearby positions.
\begin{itemize}
    \item Each heap operation is $O(\log N)$.
    \item The number of swaps equals the inversion count of the permutation, which can be $\Theta(N^2)$ in the worst case (e.g., a reversed permutation).
    \item Moreover, indices may be pushed into the heap multiple times and later discarded, but the total number of heap pushes still scales with the number of performed swaps (constant-factor neighborhood updates per swap).
    \item Therefore, worst-case time complexity is $O(\mathrm{inv}(P)\log N)=O(N^2\log N)$, with $O(N)$ memory for the array and heap.
\end{itemize}

\paragraph{After: Fenwick tree (BIT) aggregation.}
The \texttt{after} solution processes values in input order and uses a Fenwick tree to count how many smaller elements have appeared so far. For each value \texttt{val}, it infers how far this element must move (in terms of adjacent swaps) in the stable insertion process, and accumulates the corresponding cost by a closed-form arithmetic progression.
\begin{itemize}
    \item Each \texttt{query} and \texttt{update} is $O(\log N)$, performed once per element.
    \item Total time complexity is $O(N\log N)$ per test case.
    \item Space complexity is $O(N)$ for the BIT and input array.
\end{itemize}

\paragraph{Conclusion.}
Compared with the swap-simulating heap method that can degrade to $O(N^2\log N)$, the BIT-based solution computes the minimum cost in $O(N\log N)$ by aggregating necessary movements and costs without explicitly simulating each adjacent swap.

\end{document}